\documentclass{article}

\usepackage[final]{nips_2018}

\usepackage[utf8]{inputenc} 
\usepackage[T1]{fontenc}    
\usepackage{hyperref}       
\usepackage{url}            
\usepackage{booktabs}       
\usepackage{amsfonts}       
\usepackage{nicefrac}       
\usepackage{microtype}      
\usepackage{floatrow}
\usepackage{subfigure}
\usepackage{todonotes} 
\usepackage{amsmath}
\usepackage{amssymb}
\usepackage{adjustbox}
\usepackage{multirow}
\usepackage{algorithm}
\usepackage[noend]{algpseudocode}
\usepackage{accents}
\usepackage{epstopdf}

\DeclareMathOperator{\tr}{tr}

\newcommand{\calO}{\mathcal{O}}

\newcommand{\bbR}{\mathbb{R}}

\newcommand{\had}{\odot}
\DeclareMathOperator{\dd}{\partial\!}
\DeclareMathOperator{\ddt}{\partial^2\!}

\title{Scaling Gaussian Process Regression with Derivatives}

\author{
    David Eriksson,
    Kun Dong,   
    Eric Hans Lee,  
    David Bindel,
    Andrew Gordon Wilson \\
    Cornell University
}

\graphicspath{{./pics/}}
\begin{document}

\maketitle

\begin{abstract}
Gaussian processes (GPs) with derivatives
are useful in many applications, including Bayesian
optimization, implicit surface reconstruction, and terrain
reconstruction.
Fitting a GP to function values and derivatives
at $n$ points in $d$ dimensions requires linear solves and log determinants
with an ${n(d+1) \times n(d+1)}$ positive definite matrix
-- leading to prohibitive $\mathcal{O}(n^3d^3)$ computations
for standard direct methods. We propose
iterative solvers using fast $\mathcal{O}(nd)$ matrix-vector
multiplications (MVMs), together with pivoted Cholesky preconditioning
that cuts the iterations to convergence by several orders of magnitude,
allowing for fast kernel learning and prediction.
Our approaches, together 
with dimensionality reduction, enables
Bayesian optimization with derivatives to scale to high-dimensional problems and
large evaluation budgets.

\end{abstract}

\section{Introduction}
\label{sec:introduction}
Gaussian processes (GPs) provide a powerful probabilistic learning framework,
including a \emph{marginal likelihood} which represents the probability of data given only
kernel hyperparameters.  The marginal likelihood automatically
balances model fit and complexity terms to favor the simplest
models that explain the data
\citep{rasmussen01, rasmussen06, wilson2015human}. Computing the model fit term, 
as well as the predictive moments of the GP, requires solving linear systems with the kernel matrix,
while the
complexity term, or \emph{Occam's
factor}~\citep{mackay2003information}, is the log determinant of the
kernel matrix. For $n$ training points, exact kernel learning costs of $\calO(n^3)$ flops
and the prediction cost of $\calO(n)$ flops per test point are
computationally infeasible for datasets with more than a few thousand points.  The situation becomes
more challenging if we consider GPs with both function value and
derivative information, in which case training and prediction become
$\calO(n^3d^3)$ and $\calO(nd)$ respectively~\citep[\S
9.4]{rasmussen06}, for $d$ input dimensions.

Derivative information is important in many applications, including
Bayesian Optimization (BO) \citep{wu2017bayesian}, implicit surface
reconstruction
\citep{macedo2011hermite},
and terrain reconstruction.  For many simulation models, derivatives
may be computed at little extra cost via finite differences, complex
step approximation, an adjoint method, or algorithmic differentiation
\citep{forrester2008engineering}.
But while many scalable approximation methods for Gaussian
process regression have been proposed, scalable methods incorporating
derivatives have received little attention.
In this paper, we propose scalable methods for GPs with derivative
information built on the {\em structured kernel interpolation} (SKI)
framework~\citep{wilson2015kernel}, which
uses local interpolation to map scattered data onto a large grid
of inducing points, enabling fast MVMs using FFTs.
As the uniform grids in SKI scale poorly to high-dimensional spaces,
we also extend the structured kernel interpolation for products (SKIP)
method, which approximates a high-dimensional
product kernel as a Hadamard product of
low rank Lanczos decompositions \citep{gardner2018product}.
Both SKI and SKIP provide fast approximate kernel MVMs,
which are a building block to solve
linear systems with the kernel matrix and to approximate
log determinants~\citep{dong2017scalable}.

\ \newpage 
The specific contributions of this paper are:
\begin{itemize}

\item We extend SKI to incorporate derivative information, enabling
  $\calO(nd)$ complexity learning and $\calO(1)$ prediction per test
  points, relying only on fast MVM with the kernel matrix.

\item We also extend SKIP, which enables scalable Gaussian
  process regression with derivatives in high-dimensional spaces
  without grids.  Our approach allows for $\calO(nd)$ MVMs.

\item We illustrate that preconditioning is critical for fast
  convergence of iterations for kernel matrices with derivatives.  A
  pivoted Cholesky preconditioner cuts the iterations to convergence
  by several orders of magnitude when applied to both SKI and SKIP
  with derivatives.

\item We illustrate the scalability of our approach
  on several examples
  including implicit surface fitting of the
  Stanford bunny, rough terrain reconstruction, and
  Bayesian optimization.

\item We show how our methods, together with active subspace techniques,
  can be used to extend Bayesian optimization to high-dimensional
  problems with large evaluation budgets.

\item Code, experiments, and figures may be reproduced at
  \url{https://github.com/ericlee0803/GP_Derivatives}.
\end{itemize}
\vspace{-1mm}

We start in \S \ref{sec:background} by introducing GPs with
derivatives and kernel approximations.  In \S \ref{sec:methods}, we
extend SKI and SKIP to handle derivative information.
In \S \ref{sec:experiments}, we show representative experiments; and
we conclude in \S \ref{sec:discussion}. The supplementary materials
provide several additional experiments and details.

\section{Background and Challenges}
\label{sec:background}
A Gaussian process (GP) is a collection of random variables, any
finite number of which are jointly Gaussian~\citep{rasmussen06};
it also defines a distribution over functions on $\bbR^d$,
$f \sim \mathcal{GP}(\mu,k)$, where
$\mu : \bbR^d \rightarrow \bbR$ is a mean field and
$k : \bbR^d \times \bbR^d \rightarrow \bbR$ is a symmetric and
positive (semi)-definite covariance kernel.
For any set of locations $X = \{x_1,\ldots,x_n\} \subset \bbR^d$,
$f_X \sim \mathcal{N}(\mu_X, K_{XX})$ where $f_X$ and $\mu_X$
represent the vectors of function values for $f$ and $\mu$ evaluated
at each of the $x_i \in X$, and $(K_{XX})_{ij} = k(x_i, x_j)$.  We
assume the observed function value vector $y \in \bbR^n$ is contaminated by
independent Gaussian noise with variance $\sigma^2$. We denote any
kernel hyperparameters by the vector $\theta$.
To be concise, we suppress the dependence of $k$ and associated
matrices on $\theta$ in our notation.
Under a Gaussian process prior depending on the
covariance hyperparameters $\theta$, the log marginal likelihood is
given by
\begin{equation}
    \label{eq:mloglik}
    \mathcal{L}(\theta|y) =
    -\frac{1}{2}\left[(y-\mu_X)^T\alpha + \log |\tilde{K}_{XX}| + n\log 2\pi\right]
\end{equation}
where $\alpha = \tilde{K}_{XX}^{-1}(y-\mu_X)$ and
$\tilde{K}_{XX} = K_{XX} + \sigma^2 I$.
The standard direct method to evaluate (\ref{eq:mloglik}) and its
derivatives with respect to the hyperparameters uses the Cholesky
factorization of $\tilde{K}_{XX}$, leading to $\calO(n^3)$ kernel
learning that does not scale beyond a few thousand points.

A popular approach to scalable GPs is to approximate the exact kernel
with a structured kernel that enables fast MVMs
\citep{quinonero2005unifying}. Several methods approximate the kernel
via {\em inducing points} $U = \{u_j\}_{j=1}^m \subset \bbR^d$; see,
e.g.\citep{quinonero2005unifying, le2013fastfood, hensman2013uai}.
Common examples are the subset of regressors (SoR), which exploits low-rank structure,
and fully independent training conditional (FITC), which introduces an additional diagonal
correction \citep{snelson2006sparse}.
For most inducing point methods, the cost of kernel learning with $n$
data points and $m$ inducing points scales as $O(m^2 n + m^3)$, which
becomes expensive as $m$ grows.
As an alternative, Wilson and Nickisch \citep{wilson2015kernel} proposed the structured kernel interpolation (SKI) approximation,
\begin{align}
    K_{XX} \approx W K_{UU} W^{T}
    \label{eqn: ski}
\end{align}
where $U$ is a uniform grid of inducing points and
$W$ is an $n$-by-$m$ matrix of interpolation weights;
the authors of~\citep{wilson2015kernel} use local cubic interpolation
so that $W$ is sparse.
If the original kernel is stationary,
each MVM with the SKI kernel may be computed in $\calO(n + m\log(m))$
time via FFTs,
leading to
substantial performance over FITC and SoR. A limitation of SKI when used 
in combination with Kronecker inference is that
the number of grid points increases exponentially with the dimension.
This exponential scaling has been addressed by structured kernel
interpolation for products (SKIP) \citep{gardner2018product},
which decomposes the kernel matrix for a product kernel in $d$-dimensions
as a Hadamard (elementwise) product of one-dimensional kernel matrices.

We use fast MVMs to solve linear systems involving $\tilde{K}$ by
the method of conjugate gradients.  To estimate
$\log |\tilde{K}| = \tr(\log(\tilde{K}))$, we apply stochastic
trace estimators that require only products of $\log(\tilde{K})$
with random probe vectors.
Given a probe vector $z$, several ideas have been explored
to compute $\log(\tilde{K}) z$ via MVMs with $\tilde{K}$,
such as
using a polynomial approximation of $\log$ or using the connection between the Gaussian
quadrature rule and the Lanczos method \citep{han2015large,ubarufast}. It was shown in
\citep{dong2017scalable} that using Lanczos is superior to the polynomial approximations
and that only a few probe vectors are necessary even for large kernel matrices.

Differentiation is a linear operator,
and (assuming a twice-differentiable kernel) we may define a
multi-output GP for the function and (scaled) gradient values with mean and
kernel functions
\[
  \mu^{\nabla}(x) =
  \begin{bmatrix} \mu(x) \\ \partial_x \mu(x) \end{bmatrix}, \qquad
  k^{\nabla}(x,x') =
  \begin{bmatrix}
    k(x,x') & \left( \partial_{x'} k(x,x') \right)^T \\
    \partial_{x} k(x,x') & \partial^2 k(x,x')
  \end{bmatrix},
\]
where $\partial_x k(x,x')$ and $\partial^2 k(x,x')$ represent the
column vector of (scaled) partial derivatives in $x$ and the matrix
of (scaled) second partials in $x$ and $x'$, respectively.
Scaling derivatives by a natural length scale gives the multi-output
GP consistent units, and lets us understand approximation error without
weighted norms.
As in the scalar GP case,
we model measurements of the function as contaminated by independent
Gaussian noise.

\begin{figure}[!t] 
  \centering
  \includegraphics[width=0.98\textwidth]{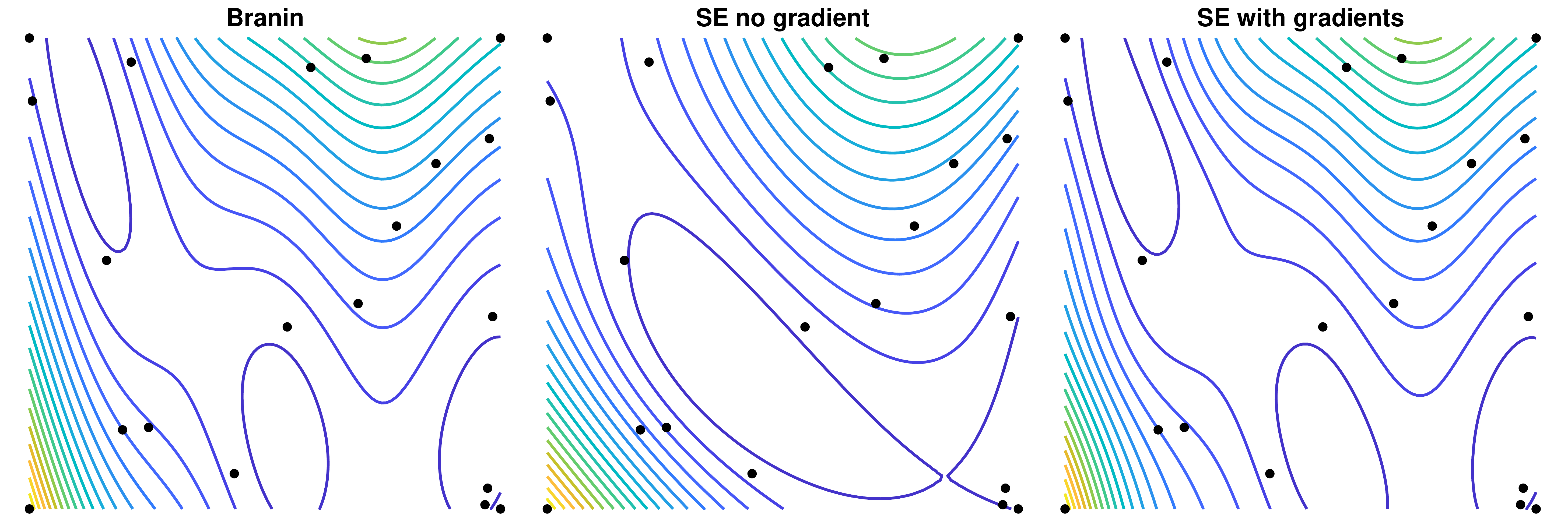}
    \caption{An example where gradient information pays off; the true function is on the left.
    Compare the regular GP without derivatives (middle) to the GP with derivatives (right). Unlike the former, the latter is able to accurately capture critical points of the function. }
    \label{fig:branin}
\end{figure}

Because the kernel matrix for the GP on function values alone is a submatrix
of the kernel matrix for function values and derivatives together,
the predictive variance in the presence of derivative information will
be strictly less than the predictive variance without derivatives.
Hence, convergence of regression with derivatives is always superior
to convergence of regression without, which is well-studied
in,~e.g.~\cite[Chapter 7]{rasmussen06}
Figure \ref{fig:branin} illustrates the value of derivative information;
fitting with derivatives is evidently much more accurate than fitting function
values alone.
In higher-dimensional problems, derivative information is
even more valuable, but it comes at a cost:
the kernel matrix $K^{\nabla}_{XX}$ is of size $n(d+1)$-by-$n(d+1)$.
Scalable approximate solvers are therefore vital in order
to use GPs for large datasets with derivative data, particularly
in high-dimensional spaces.

\section{Methods}
\label{sec:methods}
One standard approach to scaling GPs substitutes the exact kernel with an approximate kernel. 
When the GP fits values and gradients, one may attempt to separately approximate the kernel and 
the kernel derivatives. Unfortunately, this may lead to indefiniteness, as the resulting
approximation is no longer a valid kernel. Instead, we differentiate the approximate kernel,
which preserves positive definiteness. We do this for the SKI and SKIP kernels below, but our 
general approach applies to any differentiable approximate MVM.

\subsection{D-SKI}
D-SKI (SKI with derivatives) is the standard kernel
matrix for GPs with derivatives, but applied to the SKI
kernel.  Equivalently, we differentiate the interpolation scheme:
\[
k(x,x') \approx \sum_i w_i(x)k(x_i,x')
\rightarrow
\nabla k(x,x') \approx \sum_i \nabla w_i(x) k(x_i,x').
\]
One can use
cubic convolutional interpolation \citep{keys1981cubic}, but
higher order methods lead to greater accuracy, and we therefore use
quintic interpolation~\cite{meijering1999image}.  The resulting D-SKI kernel
matrix has the form
\[
    \begin{bmatrix} K & (\dd K)^T \\ \dd K & \ddt K \end{bmatrix}  \approx
    \begin{bmatrix} W \\ \dd W  \end{bmatrix}  K_{UU}
    \begin{bmatrix} W \\ \dd W  \end{bmatrix} ^T =
    \begin{bmatrix} WK_{UU}W^T & W K_{UU} (\dd W)^T \\
        (\dd W) K_{UU} W^T & (\dd W) K_{UU} (\dd W)^T  \end{bmatrix} ,
\]
where the elements of sparse matrices $W$ and  $\dd W$ are determined by $w_i(x)$
and $\nabla w_i(x)$ ---
assuming quintic interpolation, $W$ and $\dd W$ will each have $6^d$
elements per row.
As with SKI,
we use FFTs to obtain $\mathcal{O}(m \log m)$ MVMs with $K_{UU}$.
Because $W$ and $\dd W$ have $\mathcal{O}(n6^d)$ and $\mathcal{O}(nd6^d)$ 
nonzero elements, respectively, our MVM complexity is
$\mathcal{O}(nd6^d + m \log m)$.

\subsection{D-SKIP}
Several common kernels are {\em separable}, i.e., they can be expressed
as products of one-dimensional kernels.  Assuming a compatible
approximation scheme, this structure is inherited by the SKI
approximation for the kernel matrix without derivatives,
\[
    K \approx (W_1 K_1 W_1^T) \had (W_2 K_2 W_2^T) \had \ldots \had (W_d K_d W_d^T),
\]
where $A \had B$ denotes the Hadamard product of
matrices $A$ and $B$ with the same dimensions, and $W_j$ and $K_j$
denote the SKI interpolation and inducing point grid matrices
in the $j$th coordinate direction. The same Hadamard
product structure applies to the kernel matrix with derivatives;
for example, for $d = 2$,
\begin{equation} \label{skip}
\resizebox{.9\hsize}{!}{
$K^{\nabla} \approx
    \begin{bmatrix}
        W_1 K_1 W_1^T      & W_1 K_1 \dd W_1^T     & W_1 K_1 W_1^T \\
        \dd W_1 K_1 W_1^T  & \dd W_1 K_1 \dd W_1^T & \dd W_1 K_1 W_1^T \\
        W_1 K_1 W_1^T      & W_1 K_1 \dd W_1^T     & W_1 K_1 W_1^T \\
    \end{bmatrix}
    \had
    \begin{bmatrix}
        W_2 K_2 W_2^T      & W_2 K_2 W_2^T      & W_2 K_2 \dd W_2^T \\
        W_2 K_2 W_2^T      & W_2 K_2 W_2^T      & W_2 K_2 \dd W_2^T \\
        \dd W_2 K_2 W_2^T  & \dd W_2 K_2 W_2^T  & \dd W_2 K_2 \dd W_2^T \\
    \end{bmatrix} $}.
\end{equation}

Equation \ref{skip} expresses $K^{\nabla}$ as a Hadamard product of
one dimensional kernel matrices. Following this approximation, we
apply the SKIP reduction~\citep{gardner2018product} and use Lanczos to
further approximate equation \ref{skip} as $(Q_1 T_1 Q_1^T) \had (Q_2
T_2 Q_2^T)$. This can be used for fast MVMs with the kernel matrix.
Applied to kernel matrices with derivatives, we call this approach D-SKIP.

Constructing the D-SKIP kernel costs 
$\mathcal{O}(d^2(n + m \log m+ r^3 n \log d))$, and each
MVM costs $\mathcal{O}( d r^2 n)$ flops where $r$ is the effective rank of the
kernel at each step (rank of the Lanczos decomposition).
We achieve high accuracy with $r \ll n$. 

\subsection{Preconditioning}
Recent work has explored several preconditioners for exact kernel
matrices without derivatives~\citep{cutajar2016preconditioning}.
We have had success with preconditioners of the form
$M = \sigma^2 I + FF^T$ where $K^\nabla \approx FF^T$ with
$F \in \bbR^{n \times m}$.
Solving with
the Sherman-Morrison-Woodbury formula
({\em a.k.a} the matrix inversion lemma)
is inaccurate for small $\sigma$;
we use the more stable formula
$M^{-1} b = \sigma^{-2} (f-Q_1 (Q_1^T f))$
where $Q_1$ is computed in $\mathcal{O}(m^2n)$ time by
the economy QR factorization
\[
  \begin{bmatrix} Q_1 \\ Q_2 \end{bmatrix} =
  \begin{bmatrix} F \\ \sigma I \end{bmatrix} R.
\]
In our experiments with solvers for D-SKI and D-SKIP, we have found that
a truncated pivoted
Cholesky factorization,
$K^{\nabla} \approx (\Pi L)(\Pi L)^T$
works well for the low-rank factorization.
Computing the pivoted Cholesky factorization is cheaper
than MVM-based preconditioners such as Lanczos or truncated
eigendecompositions as it only requires the diagonal and the ability
to form the rows where pivots are selected. Pivoted Cholesky is a
natural choice when inducing point methods are applied as the pivoting
can itself be viewed as an inducing point method where the most
important information is selected to construct a low-rank
preconditioner~\cite{harbrecht2012low}.
The D-SKI diagonal can be formed in $\calO(nd6^d)$ flops
while rows cost $\calO(nd6^d + m)$ flops; for D-SKIP both the
diagonal and the rows can be formed in $\calO(nd)$ flops.

\subsection{Dimensionality reduction}
In many high-dimensional function approximation problems, only a few
directions are relevant.  That is, if $f : \bbR^D \rightarrow \bbR$ is
a function to be approximated, there is often a matrix $P$ with
$d < D$ orthonormal columns spanning an {\em active subspace}
of $\bbR^D$ such that $f(x) \approx f(PP^T x)$ for all $x$ in some
domain $\Omega$ of interest~\citep{constantine2015active}.  The
optimal subspace
is given by the dominant eigenvectors of the covariance matrix
$C = \int_\Omega \nabla f(x) \, \nabla f(x)^T \, dx$, 
generally estimated by Monte Carlo integration.
Once the subspace is determined, the function can be
approximated through a GP on the reduced space, i.e.,
we replace the original kernel $k(x,x')$ with a new
kernel $\check k(x,x') = k(P^T x,P^T x')$.
Because we assume gradient
information, dimensionality reduction based on active
subspaces is a natural pre-processing phase before applying D-SKI and D-SKIP.

\section{Experiments}
\label{sec:experiments}
\begin{figure}[!ht]
    \centering
    \includegraphics[width=0.98\textwidth]{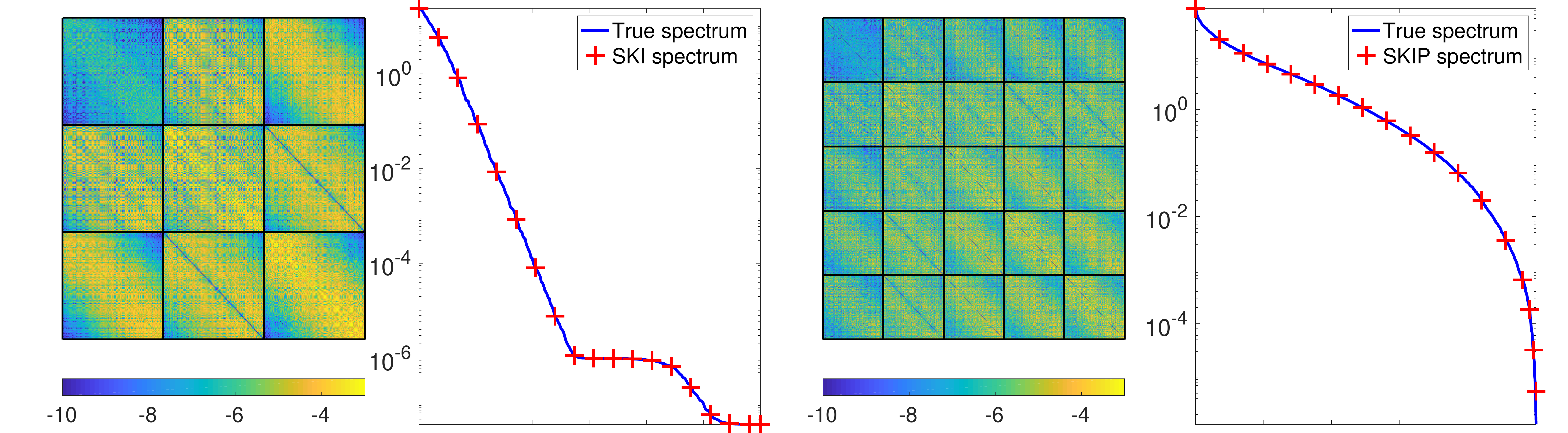}
    \caption{
      (Left two images) $\log_{10}$ error in SKI approximation and
      comparison to the exact spectrum.
      (Right two images) 
        $\log_{10}$ error in SKIP approximation and comparison to the
      exact spectrum.
    }
    \label{fig:error_ski}
\end{figure}

Our experiments use the squared exponential (SE) kernel, which has
product structure and can be used with D-SKIP; and the
spline kernel, to which D-SKIP does not directly apply.
We use these kernels in tandem with D-SKI and D-SKIP
to achieve the fast MVMs derived in \S \ref{sec:methods}.
We write D-SE to denote the exact SE kernel with derivatives.

\begin{figure}[!ht]
  {\centering \includegraphics[width=7cm]{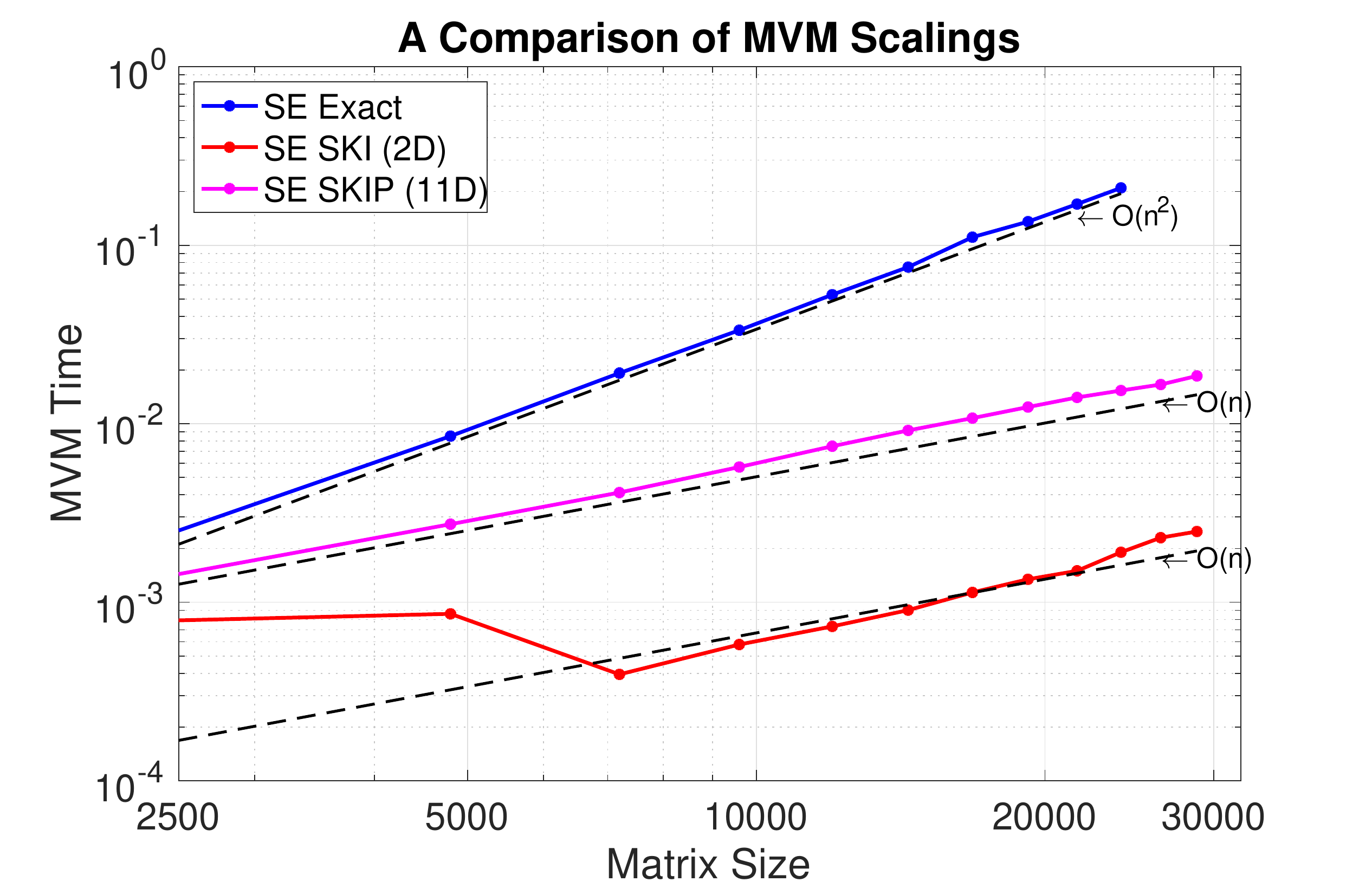}}
  \caption{Scaling tests for D-SKI in two dimensions and D-SKIP in 11 dimensions.
    D-SKIP uses fewer data points for identical matrix sizes.}
\label{fig:scalingmvm}  
\end{figure}

D-SKI and D-SKIP with the SE kernel approximate the original kernel well,
both in terms of MVM accuracy and spectral profile. Comparing D-SKI
and D-SKIP to their exact counterparts in Figure \ref{fig:error_ski},
we see their matrix entries are very close (leading to MVM accuracy
near $10^{-5}$), and their spectral profiles are indistinguishable.
The same is true with the spline kernel. 
Additionally, scaling tests in Figure \ref{fig:scalingmvm} 
verify the predicted
complexity of D-SKI and D-SKIP.
We show the relative fitting accuracy of SE, SKI, D-SE, and D-SKI
on some standard test functions in Table~\ref{fig:testfncSKI}.

\begin{table}[!ht]
\centering
 \begin{tabular}{|c|c|c|c|c|c|c|} 
 \hline
   & Branin & Franke & Sine Norm & Sixhump & StyTang & Hart3 \\ 
 \hline
 SE        & 6.02e-3 & 8.73e-3 & 8.64e-3 & 6.44e-3 & 4.49e-3 & 1.30e-2 \\
 SKI       & 3.97e-3 & 5.51e-3 & 5.37e-3 & 5.11e-3 & 2.25e-3 & 8.59e-3 \\
 D-SE   & 1.83e-3 & 1.59e-3 & 3.33e-3 & 1.05e-3 & 1.00e-3 & 3.17e-3 \\
 D-SKI     & 1.03e-3 & 4.06e-4 & 1.32e-3 & 5.66e-4 & 5.22e-4 & 1.67e-3 \\ 
 \hline
\end{tabular}
\caption{
Relative RMSE error on 10000 testing points for test functions
from~\cite{sfutest2013}, including five 2D functions (Branin,
Franke, Sine Norm, Sixhump, and Styblinski-Tang) and the 3D Hartman function.
We train the SE kernel on $4000$ points, 
the D-SE kernel on $4000/(d+1)$ points,
and SKI and D-SKI with SE kernel on $10000$
points to achieve comparable runtimes between
methods.
}
\label{fig:testfncSKI}
\end{table}

\subsection{Dimensionality reduction}

\begin{figure}[!ht]
    \centering
    \subfigure[\scriptsize Log Directional Variation]{\label{fig:dir_var}
    \includegraphics[width=0.237\textwidth,trim=1cm .5cm 2.5cm 1.5cm,clip]{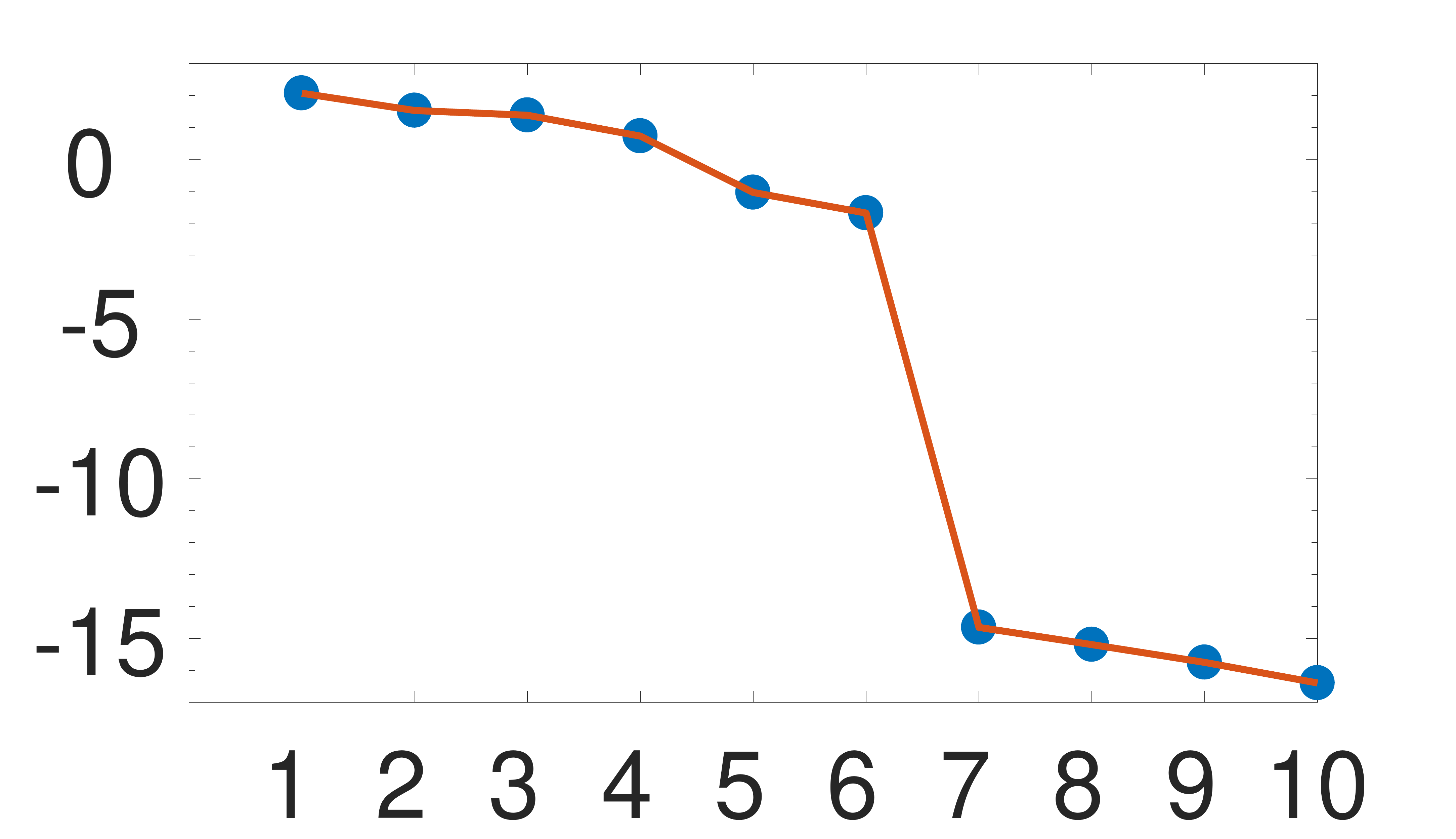}} 
    \subfigure[\scriptsize First Active Direction]{\label{fig:d1_sca}
    \includegraphics[width=0.237\textwidth,trim=1cm .5cm 2.5cm 1.5cm,clip]{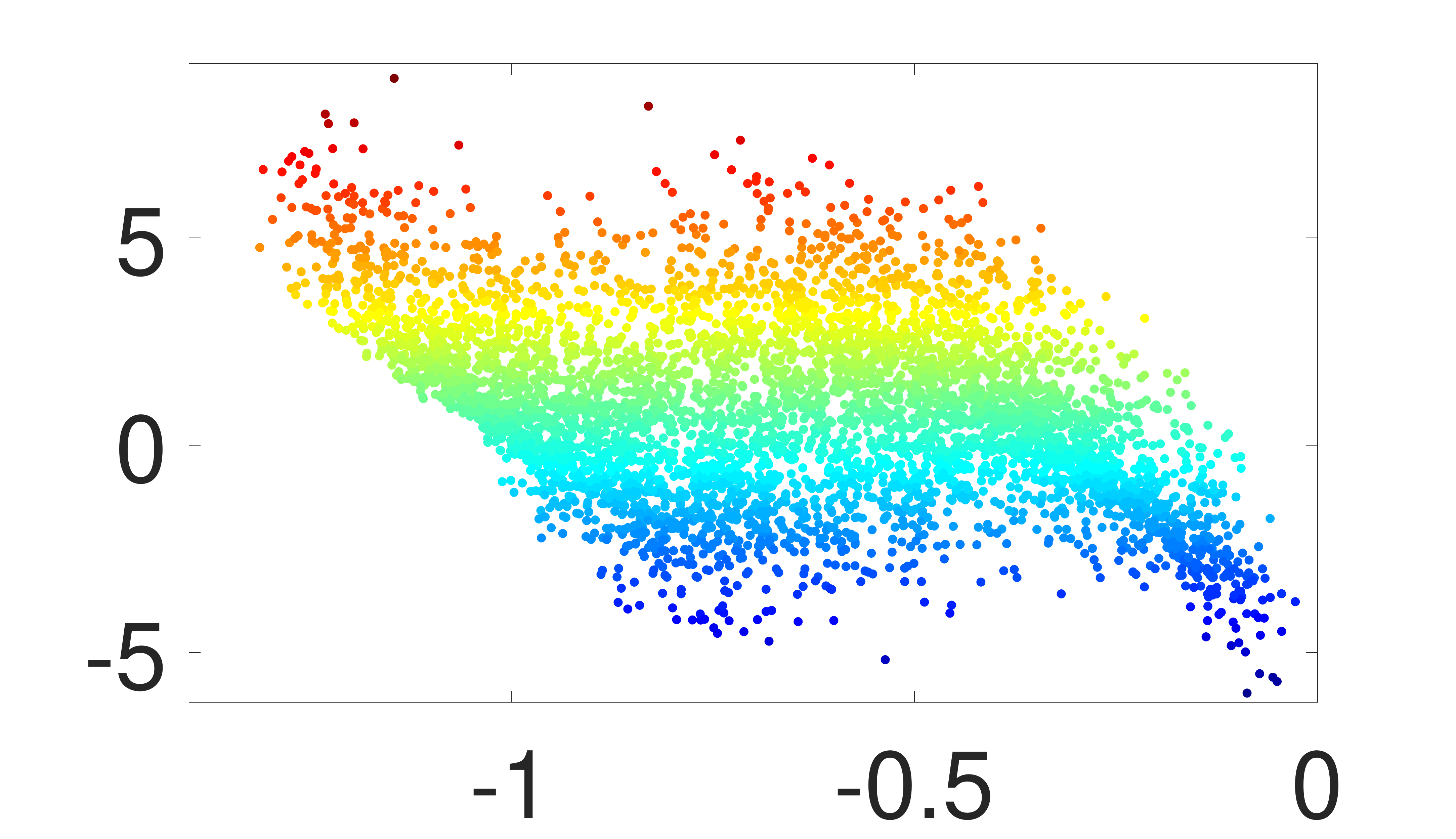}} 
    \subfigure[\scriptsize Second Active Direction]{\label{fig:d2_sca}
    \includegraphics[width=0.237\textwidth,trim=1cm .5cm 2.5cm 1.5cm,clip]{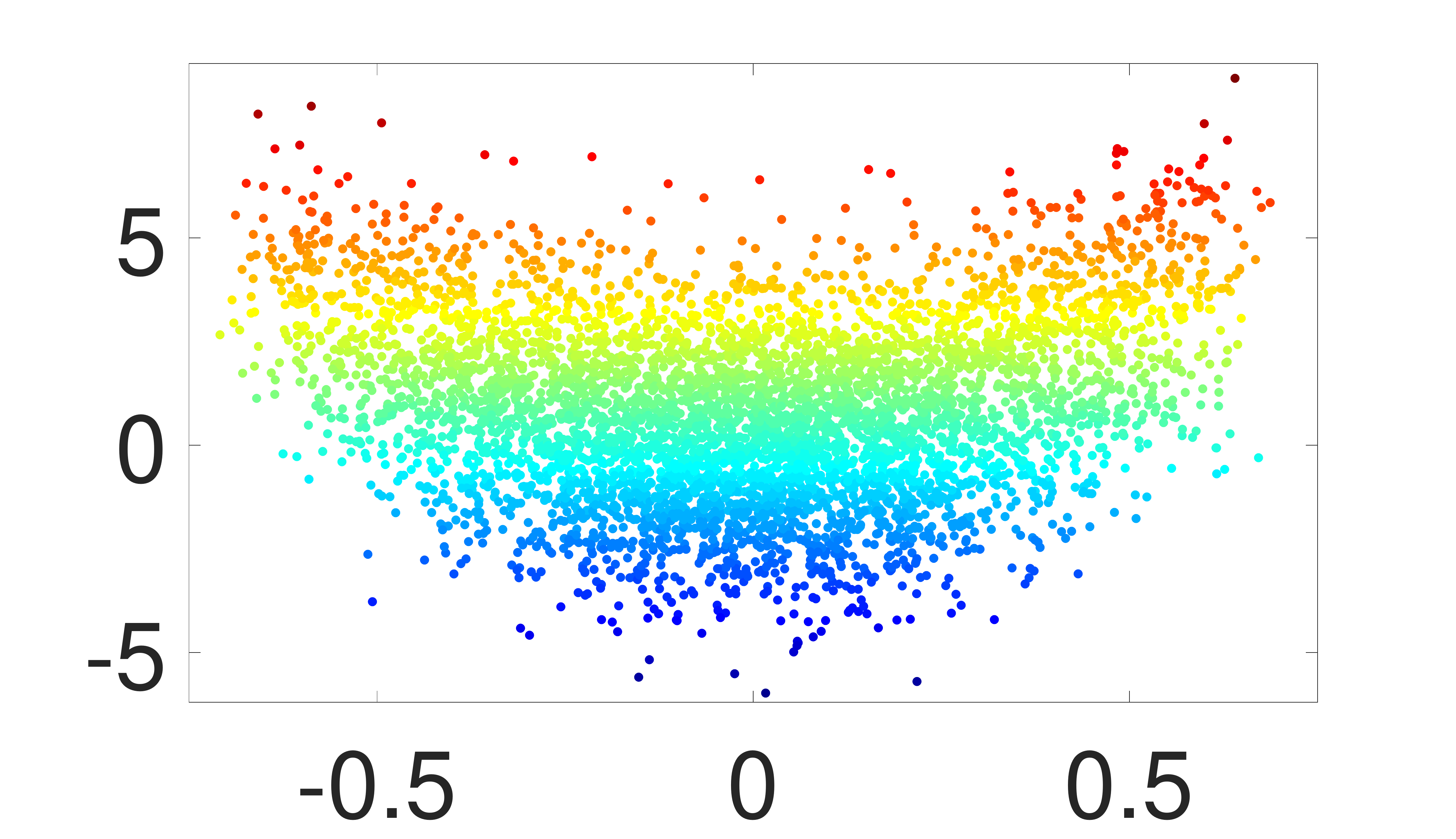}} 
    \subfigure[\scriptsize Leading 2D Active Subspace]{\label{fig:joint_sca}
    \includegraphics[width=0.237\textwidth,trim=1cm .5cm 2.5cm 1.5cm,clip]{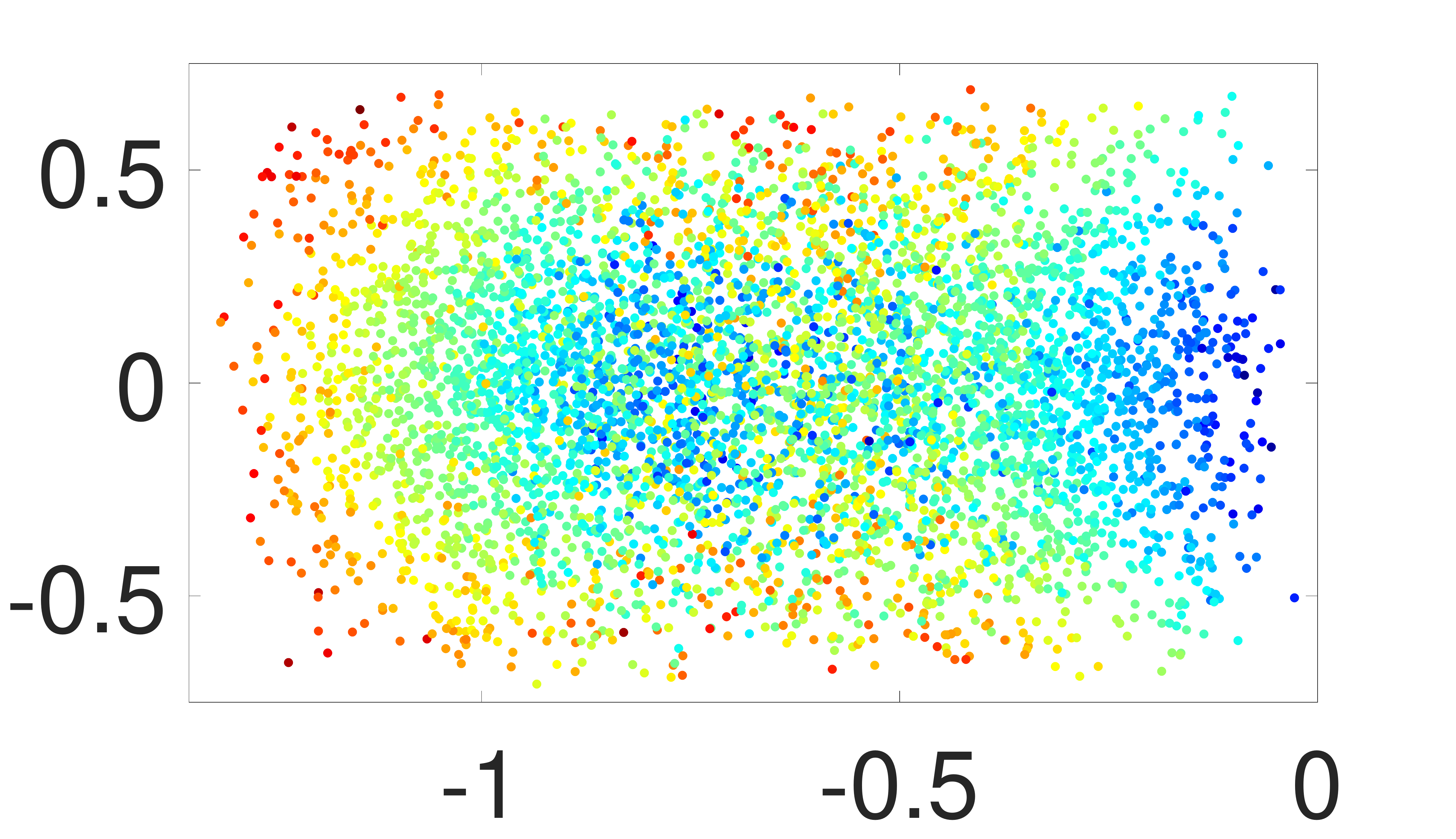}} 
\caption{\ref{fig:dir_var} shows the top $10$ eigenvalues of the gradient covariance. 
Welsh is projected onto the first and second active direction in \ref{fig:d1_sca} and 
\ref{fig:d2_sca}. After joining them together, we see in \ref{fig:joint_sca} that points 
of different color are highly mixed, indicating a very spiky surface.} 
\label{fig:dim_red}
\end{figure}

We apply active subspace pre-processing to the 20 dimensional Welsh
test function in \cite{ben2007modeling}. The top six eigenvalues of
its gradient covariance matrix are well separated from the rest as seen 
in Figure \ref{fig:dir_var}. However, the function is far from smooth 
when projected onto the leading 1D or 2D active subspace, as Figure 
\ref{fig:d1_sca} - \ref{fig:joint_sca} indicates, where the color shows
the function value.

We therefore apply D-SKI and D-SKIP on the 3D and 6D active subspace,
respectively, using $5000$ training points, and compare the prediction
error against D-SE with $190$ training points because of our scaling
advantage. Table \ref{tab:dim_red} reveals that while the 3D active
subspace fails to capture all the variation of the
function, the 6D active subspace is able to do so.
These properties are demonstrated by the poor prediction of D-SKI in
3D and the excellent prediction of D-SKIP in 6D. 

\begin{table}[!ht]
\centering
 \begin{tabular}{|c|c|c|c|} 
 \hline
   & D-SE & D-SKI (3D) & D-SKIP (6D) \\ 
 \hline
 RMSE      & 4.900e-02 & 2.267e-01 & 3.366e-03 \\
 SMAE      & 4.624e-02 & 2.073e-01 & 2.590e-03 \\
 \hline
\end{tabular}
\caption{
  Relative RMSE and SMAE prediction error for Welsh.
  The D-SE kernel is trained on $4000/(d+1)$ points, with D-SKI and
  D-SKIP trained on $5000$ points. The 6D active subspace is sufficient 
  to capture the variation of the test function.
}
\label{tab:dim_red}
\end{table}

\subsection{Rough terrain reconstruction}
Rough terrain reconstruction is a key application in
robotics \cite{gingras2010rough, konolige2010large}, autonomous
navigation \cite{hadsell2010accurate}, and geostatistics. Through a
set of terrain measurements, the problem is to predict the
underlying topography of some region.  In the following experiment,
we consider roughly $23$ million non-uniformly sampled elevation
measurements of Mount St. Helens obtained via
LiDAR \cite{sthelen2002lidar}. We bin the measurements into a
$970\times 950$ grid, and downsample to a $120\times 117$ grid.
Derivatives are approximated using a finite difference scheme.

\begin{figure}[!ht]
    \centering
    \includegraphics[width = \textwidth, height = 0.47\textwidth,
      trim = 0cm 0cm 0cm 0cm,clip]{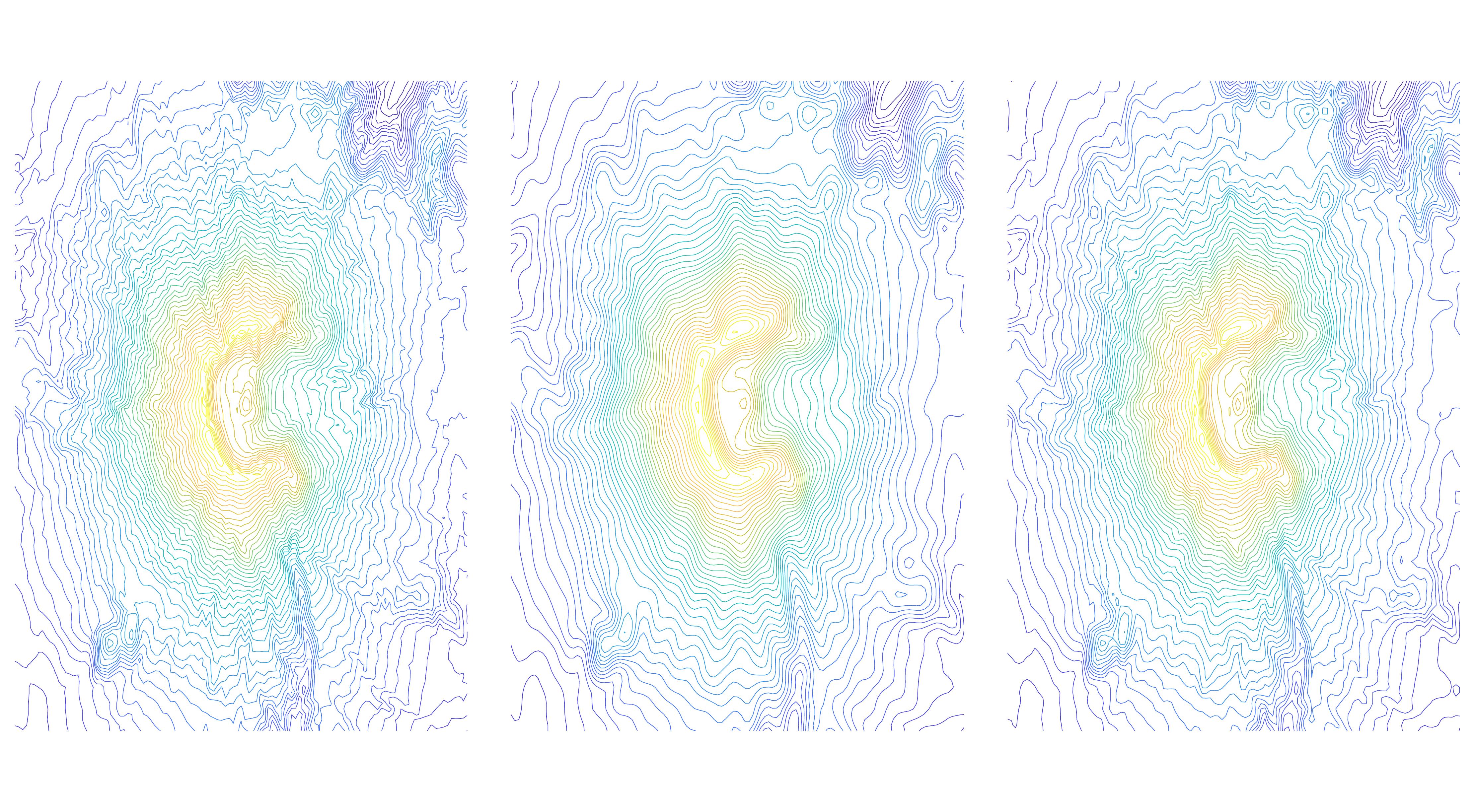} 
    \caption{On the left is the true elevation map of Mount St. Helens. In the middle is 
    the elevation map calculated with the SKI. On the right is the elevation
    map calculated with D-SKI. }
    \label{fig:MtSH_map}
\end{figure}

We randomly select $90\%$ of the grid for training and the remainder
for testing. We do not include results for D-SE, as its kernel
matrix has dimension roughly $4 \cdot 10^4$. We plot contour maps
predicted by SKI and D-SKI in
Figure \ref{fig:MtSH_map} ---the latter looks far closer to the ground
truth than the former. This is quantified in
the following table:

\begin{table}[ht]
\centering
\begin{tabular}{|c|c|c|c|c|c|c|c|}
\hline
& $\ell$ & $s$ & $\sigma$    & $\sigma_2$ & Testing SMAE & Overall SMAE& Time[s]\\ \hline
SKI & $35.196$ & $207.689$   & $12.865$ & n.a. & $0.0308$ &  $0.0357 $ & $37.67$\\ \hline
D-SKI & $12.630$ & $317.825$ & $6.446$ & $2.799$ & $0.0165$ & $0.0254 $ & $131.70$\\ \hline
\end{tabular}
\caption{The hyperparameters of SKI and D-SKI are listed. Note that there are two different
noise parameters $\sigma_1$ and $\sigma_2$ in D-SKI, for the value and gradient respectively.}
\end{table}

\subsection{Implicit surface reconstruction}
Reconstructing surfaces from point cloud data and surface
normals is a standard problem in computer vision and graphics.
One popular approach is to fit an implicit function
that is zero on the surface with gradients equal to the surface normal.
Local Hermite RBF interpolation has been considered
in prior work \cite{macedo2011hermite},
but this approach is sensitive to noise. In our experiments, using a
GP instead of splining
reproduces implicit
surfaces with very high accuracy.  In this case, a GP with derivative
information is required, as the function values are all zero.

\begin{figure}[!ht]
    \centering
    \includegraphics[width=0.9\textwidth]{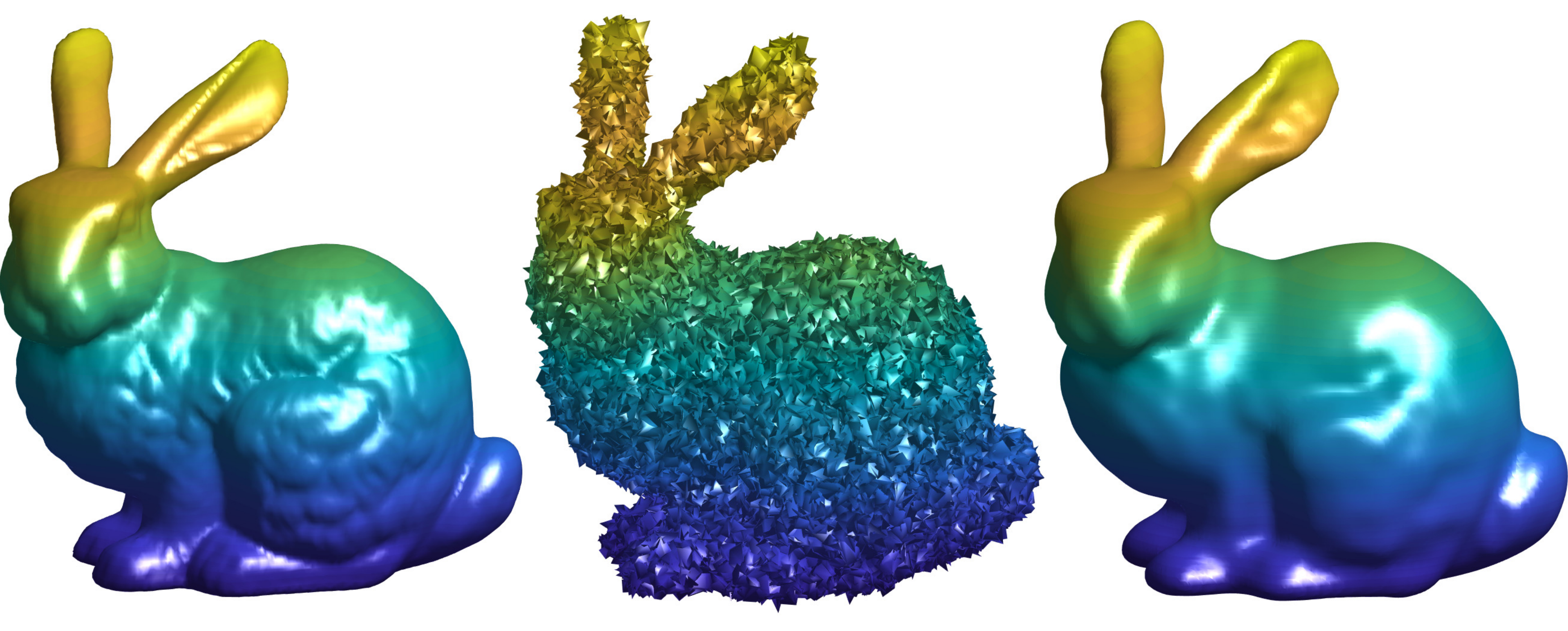}
    \caption{(Left) Original surface (Middle) Noisy surface (Right) SKI 
        reconstruction from noisy surface ($s=0.4, \sigma=0.12$)}
    \label{fig:bunny}
\end{figure}

In Figure \ref{fig:bunny}, we fit the Stanford bunny using $25000$
points and associated normals, leading to a $K^{\nabla}$ matrix of
dimension $10^5$, clearly far too large for exact training. We
therefore use SKI with the thin-plate spline kernel, with a total of
30 grid points in each dimension. The left image is a ground truth
mesh of the underlying point cloud and normals.  The middle image
shows the same mesh, but with heavily noised points and normals. Using
this noisy data, we fit a GP and reconstruct a surface shown in the
right image, which looks very close to the original.

\subsection{Bayesian optimization with derivatives}

Prior work examines Bayesian optimization (BO) with derivative
information in low-dimensional spaces to optimize model
hyperparameters \citep{wu2017bayesian}.  Wang et al.~consider
high-dimensional BO (without gradients) with random projections
uncovering low-dimensional structure~\citep{wang2013bayesian}.  We
propose BO with derivatives and dimensionality reduction via active
subspaces, detailed in Algorithm \ref{alg:BO}.

\begin{algorithm}
\caption{BO with derivatives and active subspace learning}
\label{alg:BO}
\begin{algorithmic}[1]
\While {Budget not exhausted}
  \State  Calculate active subspace projection $P \in \mathbb{R}^{D \times d}$ using sampled gradients\;
  \State  Optimize acquisition function,  $u_{n+1} = \text{arg max }\mathcal{A}(u)$ with $x_{n+1} = Pu_{n+1}$
  \State  Sample point $x_{n+1}$, value $f_{n+1}$, and gradient $\nabla f_{n+1}$ 
  \State  Update data $\mathcal{D}_{i+1} = \mathcal{D}_i \cup \{ x_{n+1}, f_{n+1}, \nabla f_{n+1}\}$
  \State  Update hyperparameters of GP with gradient defined by kernel $k(P^Tx, P^Tx')$
\EndWhile
\State \textbf{end}
\end{algorithmic}
\end{algorithm}

Algorithm 1 estimates the active subspace and fits a GP with
derivatives in the reduced space.  Kernel learning, fitting, and
optimization of the acquisition function all occur in this
low-dimensional subspace.  In our tests, we use the expected
improvement (EI) acquisition function, which involves both the
mean and predictive variance.  We consider two approaches to
rapidly evaluate the predictive variance
$v(x) = k(x,x)-K_{xX} \tilde{K}^{-1} K_{Xx}$ at a test point $x$.  In the first
approach, which provides a biased estimate of the predictive variance,
we replace $\tilde{K}^{-1}$ with the preconditioner solve computed by
pivoted Cholesky; using the stable QR-based evaluation algorithm, we have
\[
  v(x) \approx \hat{v}(x) \equiv k(x,x) - \sigma^{-2} (\|K_{Xx}\|^2-\|Q_1^T K_{Xx}\|^2).
\]
We note that the approximation $\hat{v}(x)$ is always a (small)
overestimate of the true predictive variance $v(x)$.
In the second approach, we use a randomized estimator as
in~\cite{Bekas:2007:EDM} to compute the predictive variance at many points
$X'$ simultaneously, and use the pivoted Cholesky approximation as a
control variate to reduce the estimator variance:
\[
  v_{X'} = \operatorname{diag}(K_{X'X'}) -
  \mathbb{E}_z\left[
    z \had (K_{X'X} \tilde{K}^{-1} K_{XX'} z - K_{X'X} M^{-1} K_{XX'} z)
  \right] -
  \hat{v}_{X'}.
\]
The latter approach is unbiased, but gives very noisy estimates unless
many probe vectors $z$ are used.  Both the pivoted Cholesky
approximation to the predictive variance and the randomized estimator
resulted in similar optimizer performance in our experiments.

\begin{figure}[t]
    \centering
    \subfigure[\scriptsize BO on Ackley]{\label{fig:bo_ack}
    \includegraphics[width=0.47\textwidth,trim=1.5cm 0cm 1.5cm 1cm,clip]{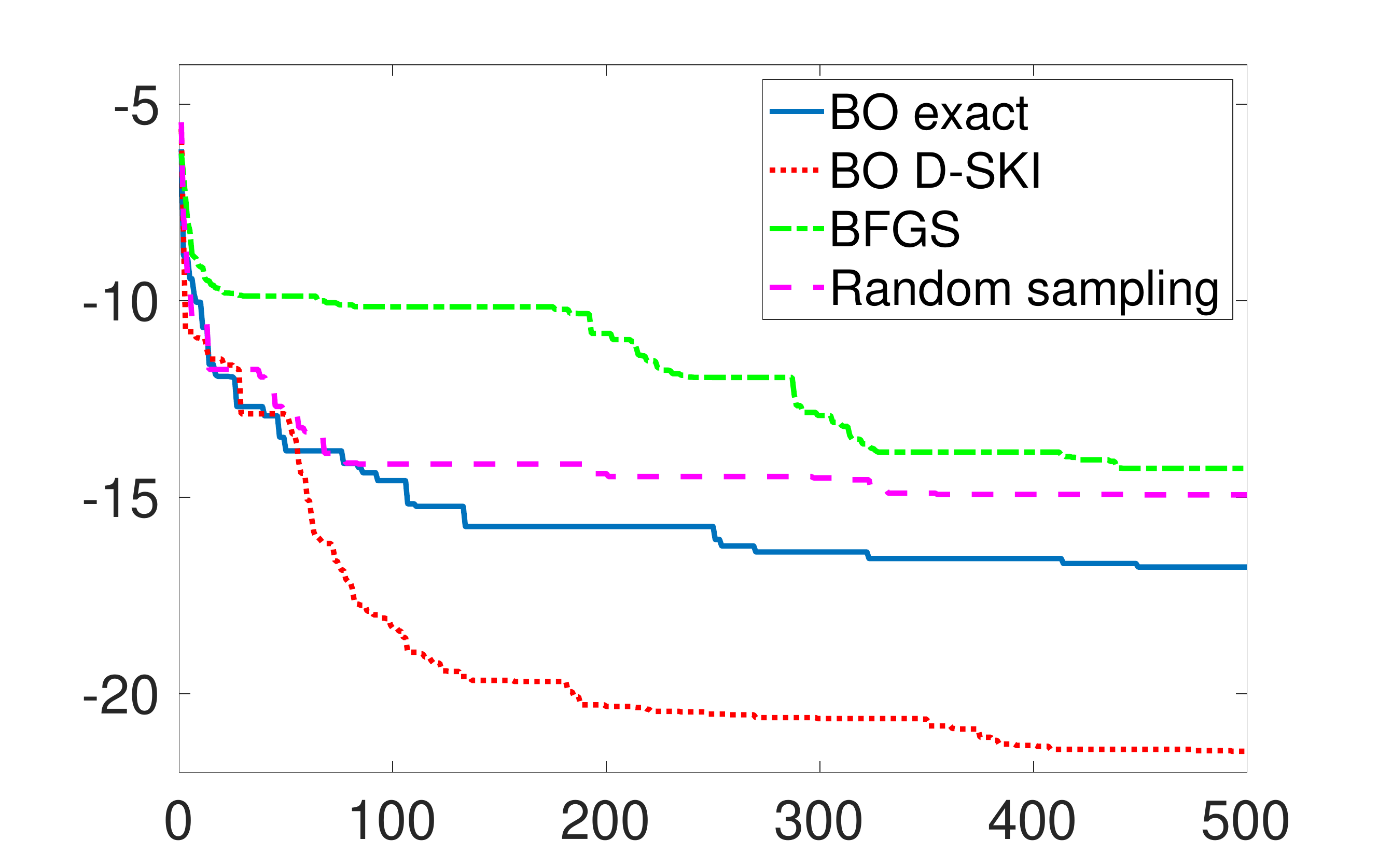}} 
    \subfigure[\scriptsize BO on Rastrigin]{\label{fig:bo_ras}
    \includegraphics[width=0.47\textwidth,trim=1.5cm 0cm 1.5cm .5cm,clip]{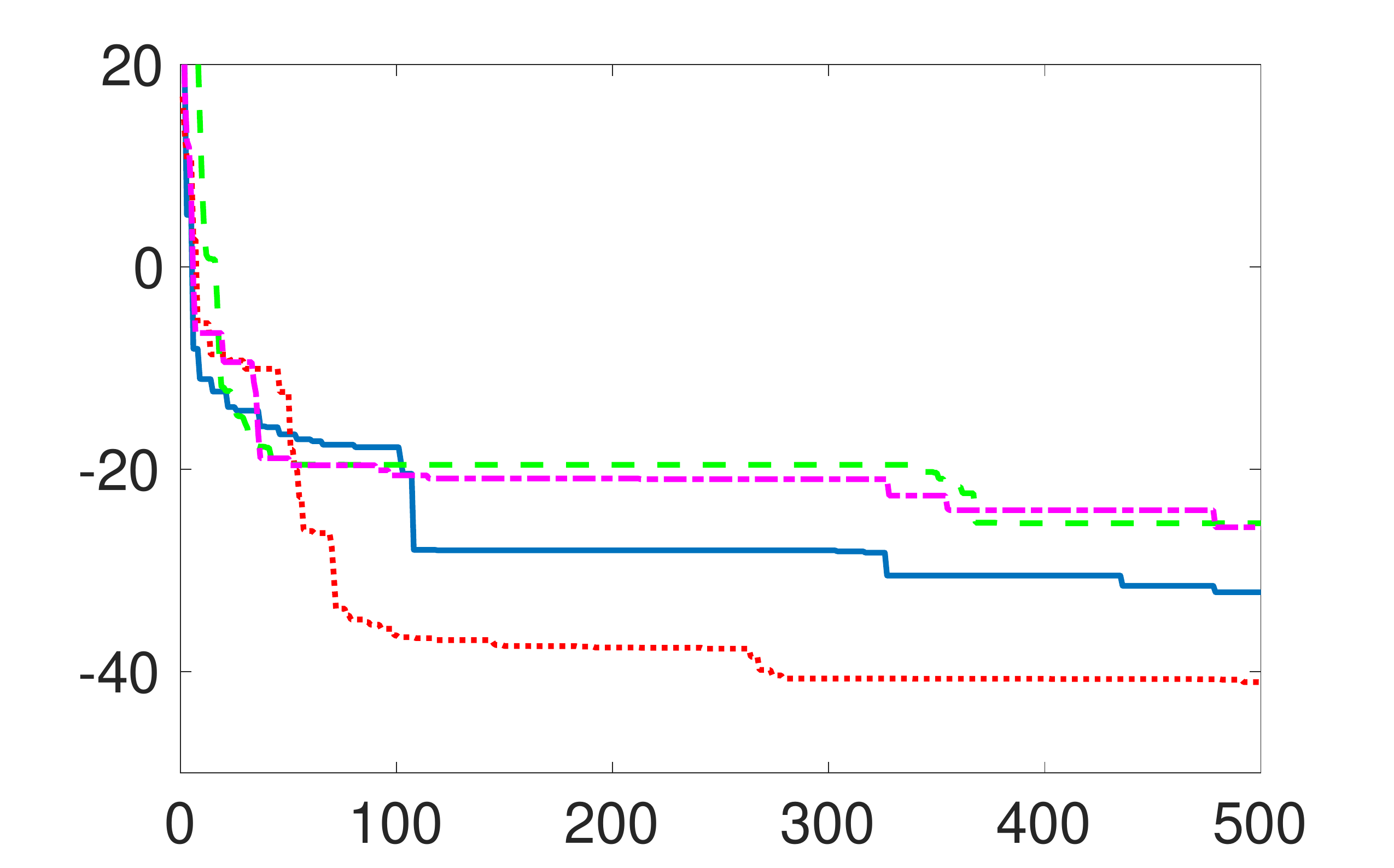}} 
\caption{In the following experiments, 5D Ackley and 5D Rastrigin are embedded into 50 
a dimensional space. We run Algorithm 1, comparing it 
with BO exact, multi-start BFGS, and random sampling. D-SKI with active subspace learning clearly outperforms the other methods.}
\end{figure}

To test Algorithm 1, we mimic the experimental set up in \cite{wang2013bayesian}:
we minimize the 5D Ackley and 5D Rastrigin test functions \cite{sfutest2013}, 
randomly embedded respectively in $[-10, 15]^{50}$ and $[-4, 5]^{50}$. 
We fix $d=2$, and at each iteration pick two directions
in the estimated active subspace at random to be our active subspace projection $P$. 
We use D-SKI as the kernel and EI as the acquisition function. 
The results of these experiments are shown in Figure \ref{fig:bo_ack} and Figure \ref{fig:bo_ras}, in which we compare Algorithm 1 
to three other baseline methods: BO with EI and no gradients in the original space;
multi-start BFGS with full gradients; and random search.  In both
experiments, the BO variants perform better than the alternatives,
and our method outperforms standard BO. 

\section{Discussion} 
\label{sec:discussion}
When gradients are available, they are a valuable source of
information for Gaussian process regression; but
inclusion of $d$ extra
pieces of information per point naturally leads to new scaling
issues.
We introduce two methods to deal with these scaling issues: D-SKI and D-SKIP.
Both are structured interpolation methods, and the
latter also uses kernel product structure. We have also discussed
practical details ---preconditioning is necessary to guarantee
convergence of iterative methods and active subspace calculation
reveals
low-dimensional structure when gradients
are available. We present several experiments with kernel
learning, dimensionality reduction, terrain reconstruction, implicit
surface fitting, and scalable Bayesian optimization with
gradients. For simplicity, these examples all possessed full gradient
information; however, our methods trivially extend if only partial
gradient information is available.

There are several possible avenues for future work.
D-SKIP shows promising scalability, but it also has large overheads,
and is expensive for Bayesian optimization as it
must be recomputed from scratch with each new data point.
We believe kernel function approximation via
Chebyshev interpolation and tensor approximation will likely
provide similar accuracy with greater efficiency.
Extracting low-dimensional structure is highly effective in our
experiments and deserves an independent, more thorough
treatment.
Finally, our work in scalable Bayesian
optimization with gradients represents a step towards the unified view
of global optimization methods (i.e.~Bayesian optimization) and
gradient-based local optimization methods (i.e.~BFGS).

\paragraph{Acknowledgements.} We thank NSF DMS-1620038, NSF IIS-1563887, and Facebook Research for support.

\bibliography{references}
\bibliographystyle{unsrtnat}

\appendix
\label{sec:supp}
\section{Kernels}
The covariance functions we consider in this paper are the squared exponential (SE) 
kernel
\[
    k_{\text{SE}}(x, y) = s^2 \exp\left(-\frac{\|x-y\|^2}{2\ell^2} \right)
\]
and the spline kernels
\begin{align*}
    k_{\text{spline}}(x, y) = 
        \begin{cases}
            s^2 \big( \| x - y \|^3 + a\| x - y \|^2 + b \big) & d \text{ odd} \\
            s^2 \big( \| x - y \|^2\,\log \| x - y \| + a\| x - y \|^2 + b \big)  & d \text{ even}
        \end{cases}
\end{align*}
where $a,b$ are chosen to make the spline kernel symmetric and positive definite on the given
domain.

\section{Kernel Derivatives}
The first and second order derivatives of the SE kernel are
\begin{align*}
    \frac{\partial k_{\text{SE}}(x^{(i)}, x^{(j)})}{\partial x_p^{(j)}} &= 
        \frac{x^{(i)}_p - x^{(j)}_p}{\ell^2} k_{\text{SE}}(x^{(i)}, x^{(j)}), \\
    \frac{\partial^2 k_{\text{SE}}(x^{(i)}, x^{(j)})}{\partial x_p^{(i)}\partial x_q^{(j)}} &= 
        \frac{1}{\ell^4}\left(\ell^2\delta_{pq} - 
        (x^{(i)}_p - x^{(j)}_p)(x^{(i)}_q - x^{(j)}_q) \right)k_{\text{SE}}(x^{(i)}, x^{(j)}).
\end{align*}
This shows that each $n$-by-$n$ block of $\partial K$ and $\partial^2$ admit Kronecker and 
Toeplitz structure if the points are on a grid.

\section{Preconditioning}
We discover that preconditioning is crucial for the convergence of iterative solvers using 
approximation schemes such as D-SKI and D-SKIP. To illustrate the performance of conjugate 
gradient (CG) method with and without the above-mentioned truncated pivoted Cholesky preconditioner, 
we test D-SKI on the 2D Franke function with $2000$ data points, and D-SKIP on the 5D Friedman 
function with $1000$ data points. In both cases, we compute a pivoted Cholesky decomposition 
truncated at rank $100$ for preconditioning, and the number of steps it takes for CG/PCG to 
converge are demonstrated in Figure \ref{fig:precond} below. It is clear that preconditioning 
universally and significantly reduces the number of steps required for convergence.

\begin{figure}[!ht]
    \centering
    \includegraphics[width=0.7\textwidth]{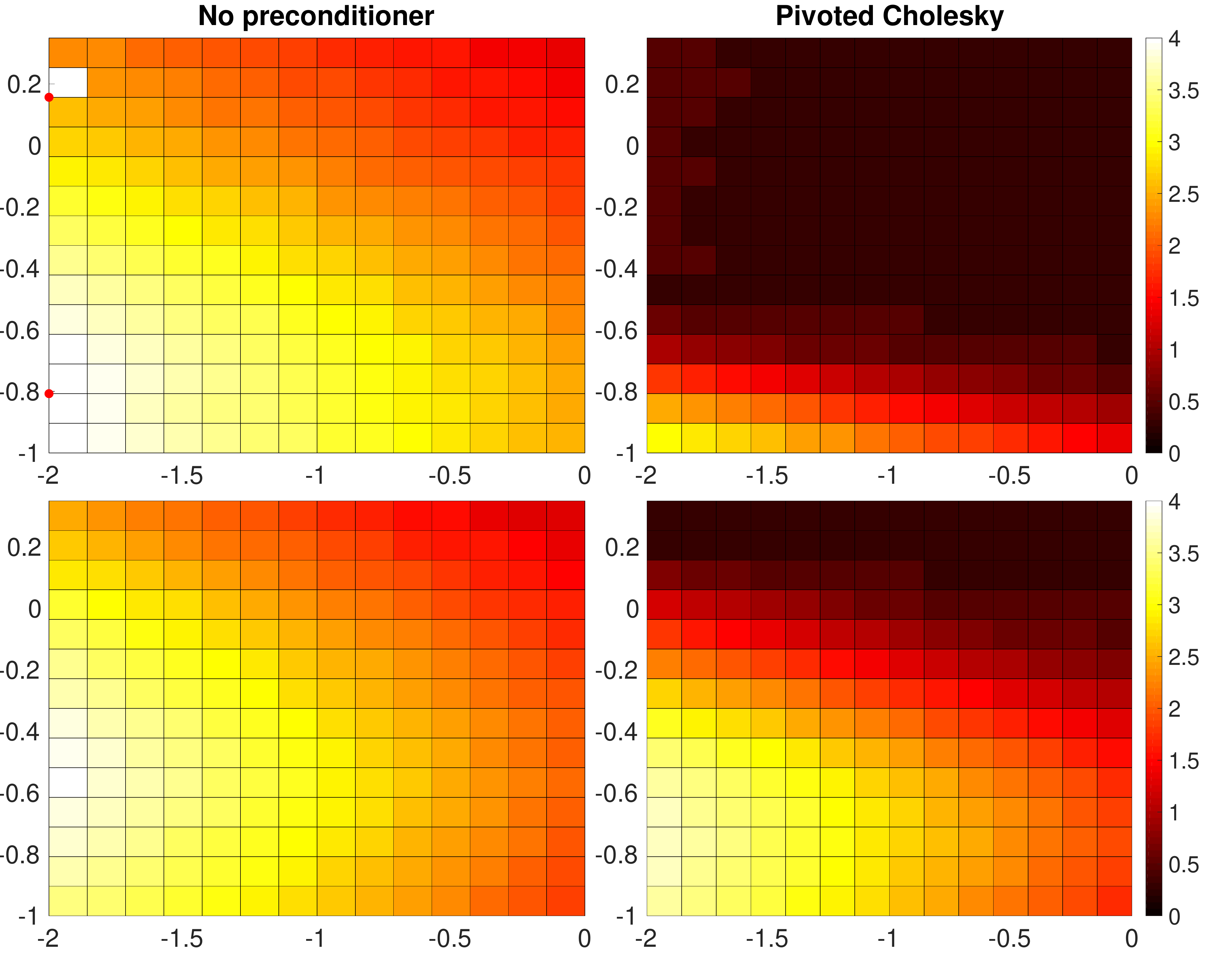}
    \caption{The color shows $\log_{10}$ of the number of iterations to reach a tolerance of 1e-4.
    The first row compares D-SKI with and without a preconditioner. The second row compares D-SKIP with and 
    without a preconditioner. The red dots represent no convergence. The y-axis shows $\log_{10}(\ell)$
    and the x-axis $\log_{10}(\sigma)$ and we used a fixed value of $s=1$.}
    \label{fig:precond}
\end{figure}

\section{Korea}
\begin{figure}[!ht]
    \centering
    \subfigure[\scriptsize Ground Truth]{\label{fig:korea_true}
    \includegraphics[width=0.30\textwidth,trim=9cm 6cm 9cm 10cm,clip]{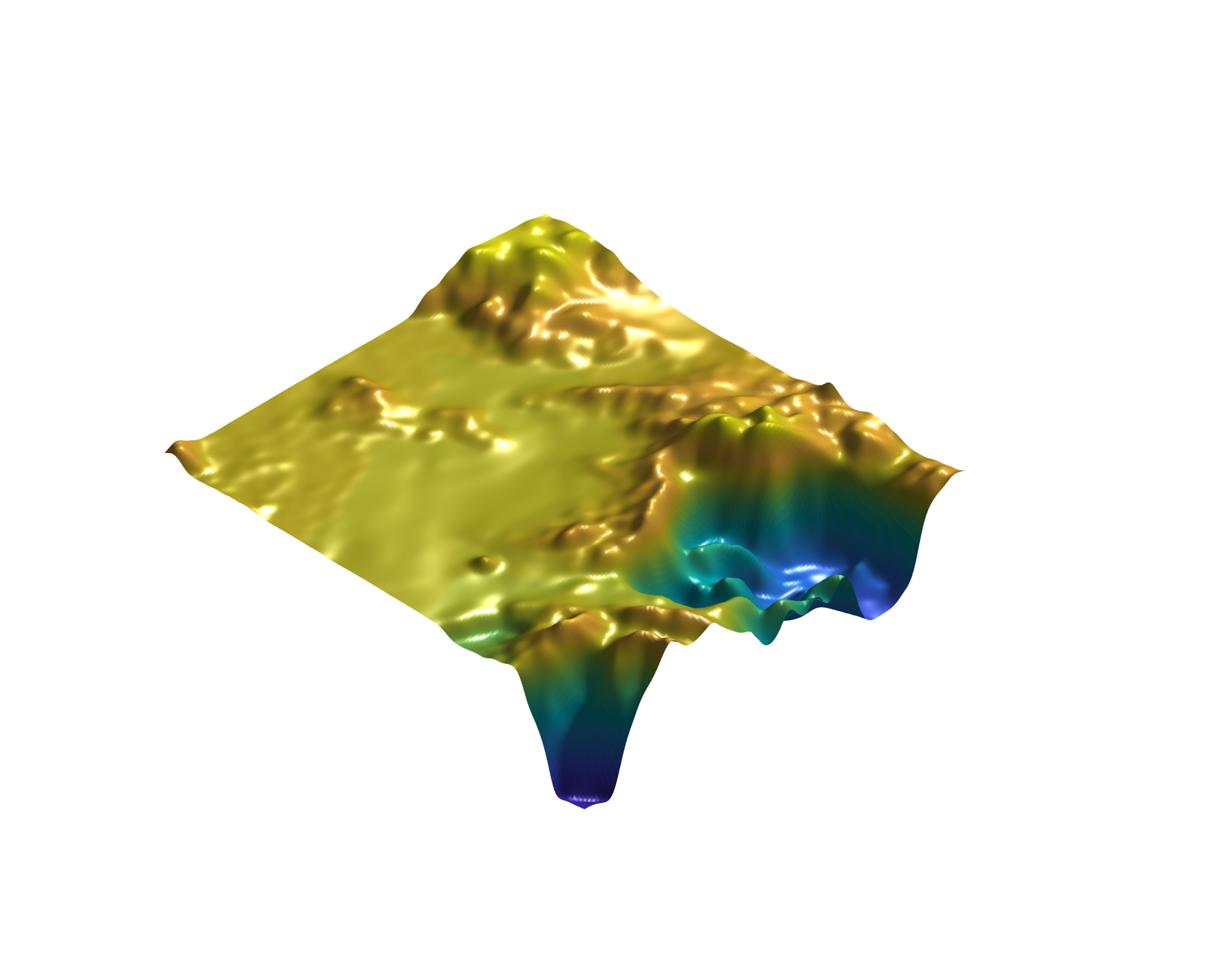}} 
    \subfigure[\scriptsize SKI]{\label{fig:korea_ski}
    \includegraphics[width=0.30\textwidth,trim=9cm 6cm 9cm 10cm,clip]{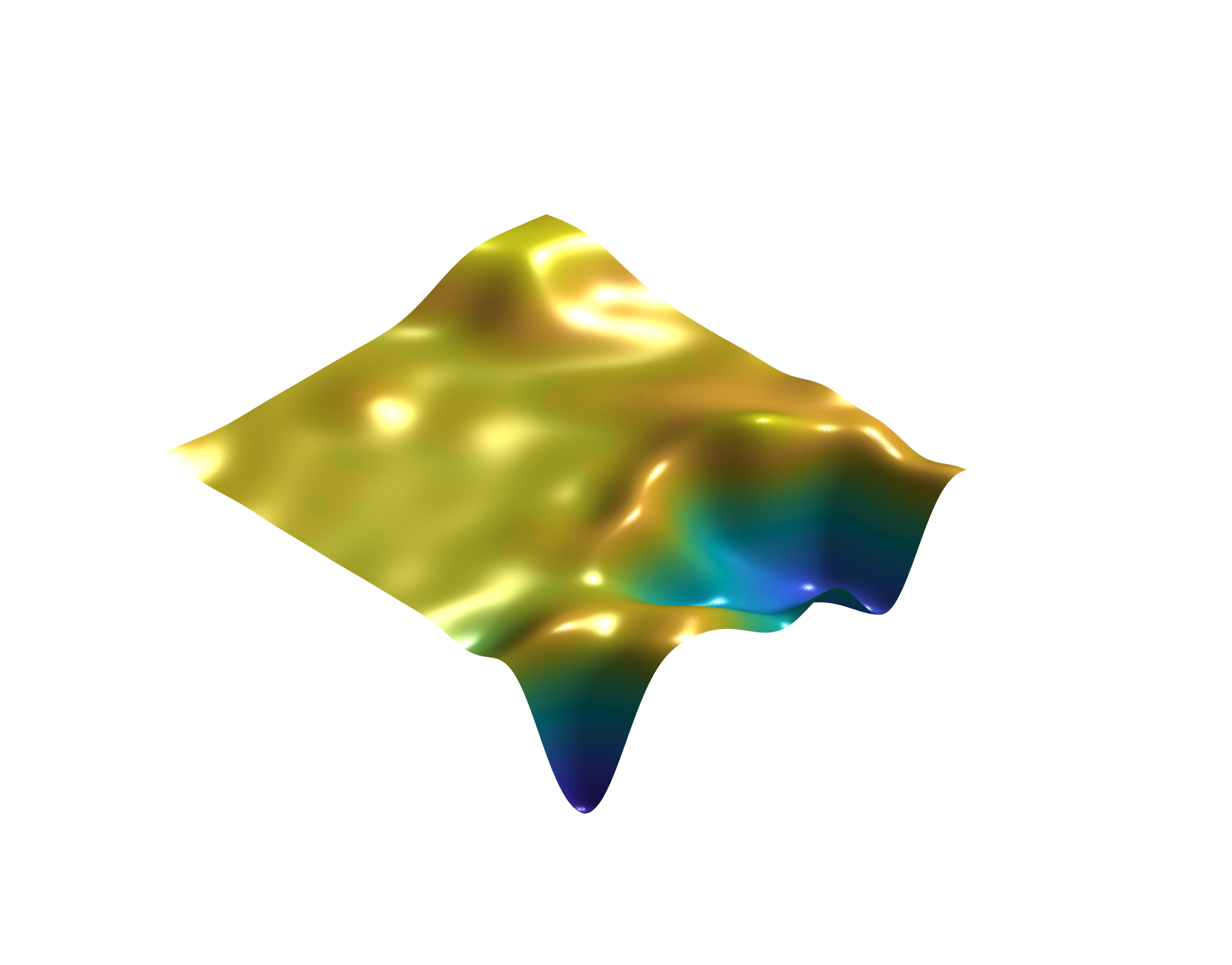}}
    \subfigure[\scriptsize D-SKI]{\label{fig:korea_ski_g}
    \includegraphics[width=0.30\textwidth,trim=9cm 6cm 9cm 10cm,clip]{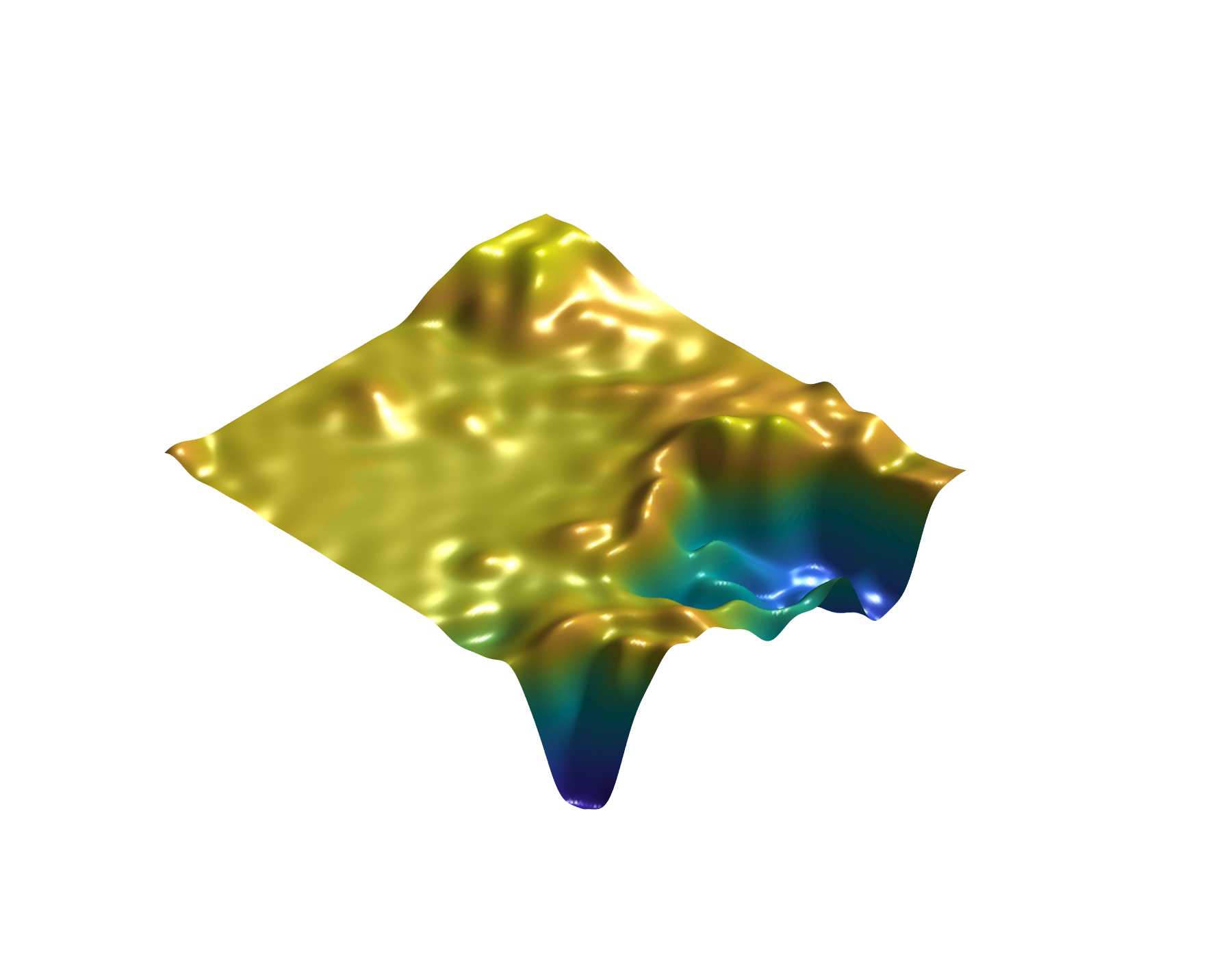}}
\caption{D-SKI is clearly able to capture more detail in the map than
  SKI.  Note that inclusion of derivative information in this case
  leads to a negligible increase in calculation time.
  } \label{fig:korea_map}
\end{figure}

The Korean Peninsula elevation and bathymetry dataset\citep{MatlabMTB} is sampled at 
a resolution of 12 cells per degree and has $180 \times 240$ entries on a rectangular
grid. We take a smaller subgrid of $17\times 23$ points as training data. To reduce data 
noise, we apply a Gaussian filter with
$\sigma_{\text{filter}} = 2$ as a pre-processing step. We observe that the 
recovered surfaces with SKI and D-SKI highly resemble their respective counterparts 
with exact computation and that incorporating gradient information enables us to recover 
more terrain detail.
\begin{table}[!ht]
    \centering
    \begin{tabular}{|c|c|c|c|c|c|}
    \hline
    & $\ell$ & $s$ & $\sigma$ & SMAE & Time[s]\\ \hline
    SKI & $16.786$ & $855.406$ & $184.253$ & $0.1521$ & $10.094$\\ \hline
    D-SKI & $9.181$ & $719.376$ & $29.486$ & $0.0746$ & $11.643$\\ \hline
    \end{tabular}
\end{table}

\end{document}